\documentclass[runningheads]{llncs}

\newif\ifcomments
\newif\ifcoverpage
\commentstrue  % to see rainbow comments
\usepackage[T1]{fontenc}
\usepackage{graphicx}
\usepackage{subcaption}
\usepackage{tikz}
\usepackage{booktabs}
\usetikzlibrary{calc}
\usepackage{amssymb}
\usepackage{listings}
\usepackage{float}
\newfloat{listing}{tbp}{lol}
\floatname{listing}{Listing}
\usepackage{hyperref}
\usepackage{siunitx}
\usepackage{algorithm}
\usepackage[noEnd=true]{algpseudocodex}
\usepackage{wrapfig}

\usepackage{amsmath}
\usetikzlibrary{arrows.meta}
\lstdefinelanguage{YAML}{
  keywords={true,false,null},
  keywordstyle=\color{black}\bfseries,
  basicstyle=\ttfamily\scriptsize,
  commentstyle=\color{gray},
  morecomment=[l]{\#},
  stringstyle=\color{black},
  moredelim=[l][\color{black}]{:},
  frame=single,
  frameround=ffff,
  framesep=3pt,
  belowcaptionskip=2pt,
  captionpos=b,
  xleftmargin=4pt,
  xrightmargin=4pt,
  columns=flexible,
  keepspaces=true,
  showstringspaces=false,
}
\usepackage{xcolor}
\usepackage{needspace}
\usepackage[table]{xcolor}
\usepackage[colorinlistoftodos,textsize=scriptsize]{todonotes}
\usepackage{xspace}
\usepackage{color}

\definecolor{blue}{RGB}{0,84,159}
\definecolor{lblue}{RGB}{64,127,183}
\definecolor{llblue}{RGB}{142,186,229}
\definecolor{lllblue}{RGB}{199,221,242}
\definecolor{llllblue}{RGB}{232,241,250}
\definecolor{black}{RGB}{0,0,0}
\definecolor{lblack}{RGB}{100,101,103}
\definecolor{llblack}{RGB}{156,158,159}
\definecolor{lllblack}{RGB}{208,209,210}
\definecolor{llllblack}{RGB}{236,237,237}

\definecolor{magenta}{RGB}{227,0,102}
\definecolor{lmagenta}{RGB}{233,96,136}
\definecolor{llmagenta}{RGB}{241,158,177}
\definecolor{lllmagenta}{RGB}{249,210,218}
\definecolor{llllmagenta}{RGB}{253,238,240}

\definecolor{yellow}{RGB}{255,237,0}
\definecolor{lyellow}{RGB}{255,240,85}
\definecolor{llyellow}{RGB}{255,245,155}
\definecolor{lllyellow}{RGB}{255,250,209}
\definecolor{llllyellow}{RGB}{255,253,,238}

\definecolor{petrol}{RGB}{0,97,101}
\definecolor{lpetrol}{RGB}{45,127,131}
\definecolor{llpetrol}{RGB}{125,164,167}
\definecolor{lllpetrol}{RGB}{191,208,209}
\definecolor{llllpetrol}{RGB}{230,236,236}

\definecolor{turquoise}{RGB}{0,152,161}
\definecolor{lturquoise}{RGB}{0,177,183}
\definecolor{llturquoise}{RGB}{137,204,207}
\definecolor{lllturquoise}{RGB}{202,231,231}
\definecolor{llllturquoise}{RGB}{235,246,246}

\definecolor{green}{RGB}{87,171,39}
\definecolor{lgreen}{RGB}{141,192,96}
\definecolor{llgreen}{RGB}{184,214,152}
\definecolor{lllgreen}{RGB}{221,235,206}
\definecolor{llllgreen}{RGB}{242,247,236}

\definecolor{grass}{RGB}{189,205,0}
\definecolor{lgrass}{RGB}{208,217,92}
\definecolor{llgrass}{RGB}{224,230,154}
\definecolor{lllgrass}{RGB}{240,43,208}
\definecolor{llllgrass}{RGB}{249,250,237}

\definecolor{orange}{RGB}{246,168,0}
\definecolor{lorange}{RGB}{250,190,80}
\definecolor{llorange}{RGB}{253,212,143}
\definecolor{lllorange}{RGB}{254,234,201}
\definecolor{llllorange}{RGB}{255,247,234}

\definecolor{red}{RGB}{204,7,30}
\definecolor{lred}{RGB}{216,92,65}
\definecolor{llred}{RGB}{230,150,121}
\definecolor{lllred}{RGB}{243,205,187}
\definecolor{llllred}{RGB}{250,235,227}

\definecolor{burgundy}{RGB}{161,16,53}
\definecolor{lburgundy}{RGB}{182,82,86}
\definecolor{llburgundy}{RGB}{205,139,135}
\definecolor{lllburgundy}{RGB}{229,197,192}
\definecolor{llllburgundy}{RGB}{245,232,229}

\definecolor{violet}{RGB}{97,33,88}
\definecolor{lviolet}{RGB}{131,78,117}
\definecolor{llviolet}{RGB}{168,133,158}
\definecolor{lllviolet}{RGB}{210,192,205}
\definecolor{llllviolet}{RGB}{237,229,234}

\definecolor{purple}{RGB}{122,111,172}
\definecolor{lpurple}{RGB}{122,111,172}
\definecolor{llpurple}{RGB}{122,111,172}
\definecolor{lllpurple}{RGB}{122,111,172}
\definecolor{llllpurple}{RGB}{122,111,172}

\definecolor{cyan}{RGB}{0,152,161}
\definecolor{lcyan}{RGB}{0,177,183}
\definecolor{llcyan}{RGB}{137,204,207}
\definecolor{lllcyan}{RGB}{202,231,231}
\definecolor{llllcyan}{RGB}{235,246,246}

\definecolor{silver}{cmyk}{.39,.31,.32,.14}
\definecolor{gold}{cmyk}{.35,.46,.7,.35}

\input{admin}
\input{acronyms}

\addauthor{wenbo}{blue}
\addauthor{daniel}{red}
\addauthor{matteo}{green}
\addauthor{till}{orange}

\altCRC{\renewcommand{\orcidID}[1]{}}{}
\extrafloats{100}

\newcommand{\algorithmFontSize}{\scriptsize}

\algrenewcommand\algorithmicindent{0.65em}
\algnewcommand\Yield{\textbf{yield }}
\algnewcommand\Continue{\textbf{continue }}
\algnewcommand\Break{\textbf{break }}

\algrenewcommand\alglinenumber[1]{\algorithmFontSize #1:}

\newcommand{\wbm}{\textsc{WorkBenchMark}\xspace}
\begin{document} 

\ifcoverpage\onecolumn
\section*{Cover Page}
\subsection*{TODOs}
\begin{itemize}
    \item Everyone: use acronyms everywhere! Add them to \texttt{acronyms.tex} if not defined yet
\end{itemize}
\subsection*{Macros}
\begin{itemize}
  \item macros are in the file \texttt{admin.tex}
  \item $\backslash$commenttill\{blah blah\} creates a comment in a color. (where till is the person commenting)
  \item You can change your color (search for \texttt{$\backslash$addauthor})
  \item $\backslash$commenttillhide\{blah blah\} hides the "blah blah"
  \item $\backslash$remove \remove{turns text light green indicating it should be removed}
  \item $\backslash$removehide hides the text
  \item $\backslash$removeifneeded \removeifneeded{highlights text in cyan that is a candidate for removal if we need space (or for some other reason).}
  \item $\backslash$revisit \revisit{turns text blue indicating it's something to reconsider/check -- good for things you're uncertain about or that someone things might be inaccurate/incorrect and warrants discussion}
  \item $\backslash$alt \alt{old text}{turns text brown and is an alternative phrasing to what preceded it}
  \item $\backslash$added \added{turns text red to highlight new text that was added --- good in final stages when people are making changes they want others to note/approve/check}
\end{itemize}

Examples:
\begin{itemize}
    \item \texttt{$\backslash$todotill}\{do something\}\todotill{do something}
    \item \texttt{$\backslash$todo[inline]}\{do something inline\}\todotill[inline]{do something inline}
    \item \texttt{$\backslash$commenttill}\{a comment\}\commenttill{a comment}
    \item \texttt{$\backslash$commenttill[noinline]}\{a noinline comment\}\commenttill[noinline]{an noinline comment}
\end{itemize}

Boolean flags are defined at the top of \texttt{paper.tex}
\begin{itemize}
    \item \texttt{$\backslash$commentstrue} enables/disables comments
    \item \texttt{$\backslash$coverpagetrue} enables/disables the cover page
\end{itemize}

\vfill
\pagebreak

\setcounter{page}{1}
\fi

%
%\title{\alt{Autonomous Robotic Assembly with Open-World Perception and Physically Feasible Planning}{\wbm: A LEGO-Based Assembly Benchmark for the RoboCup Smart Manufacturing League
%}
%\title{WorkbenchMark: A LEGO-Based Assembly Benchmark and Planning Baseline for the Smart Manufacturing League}
%\title{\wbm: A LEGO-Based Assembly Benchmark for the Smart Manufacturing League and its Assembly-by-Disassembly Baseline}
\title{\wbm: A LEGO-Based Assembly Benchmark with an Assembly-by-Disassembly Baseline for the Smart Manufacturing League}
\titlerunning{\wbm}
\author{Wenbo Ma\inst{1}\orcidID{0000-1111-2222-3333} \and
Daniel Swoboda\inst{1}\orcidID{0000-0002-3189-3089} \and
Matteo Tschesche\inst{2}\orcidID{2222--3333-4444-5555}\and
Till Hofmann\inst{1}\orcidID{2222--3333-4444-5555}
}
\authorrunning{W. Ma et al.}
\institute{Chair of Machine Learning and Reasoning (i6), RWTH Aachen University, \\
Aachen, Germany\\
\email{\{first.last\}@ml.rwth-aachen.de} \and
MASCOR Institute, FH Aachen University of Applied Sciences, Aachen, Germany\\
\email{tschesche@fh-aachen.de}}
\maketitle              % typeset the header of the contribution
\begin{abstract}
% Robotic assembly requires the integration of high-level sequencing, open-world perception, and precise manipulation under geometric and physical constraints. 
% In this paper, we introduce \wbm, a LEGO Duplo assembly benchmark inspired by the Workbench Track of the RoboCup Smart Manufacturing League. 
% The benchmark provides 400 simulation tasks across four complexity tiers, ranging from simple two-brick stacking to complex interlocking structures, together with structured YAML task specifications and a reproducible simulation environment.

% As a baseline, we present an integrated robotic system that combines Assembly-by-Disassembly (ABD) planning with open-vocabulary perception and collision-aware execution. The planner explicitly enforces grasp reachability and structural stability constraints, while perception is achieved through language-guided detection and 6D pose estimation. The resulting system produces physically feasible assembly sequences and reliable execution.

% We evaluate the proposed approach against a vision-language-action (VLA) baseline in simulation. The results show that our structured pipeline achieves higher success rates and more stable performance across task tiers, while the VLA baseline struggles with precise manipulation and constraint-aware planning in complex scenarios.

% \wbm provides a first reference point for evaluating integrated robotic assembly systems in the SML context and beyond.
We introduce \wbm, a LEGO Duplo-based robotic assembly
benchmark motivated by the RoboCup Smart Manufacturing
League. Robotic assembly couples low-level manipulation with task-level
symbolic reasoning under physical constraints, a combination that
current end-to-end learning methods do not yet solve reliably.
The benchmark provides 400 tasks across four complexity
tiers. 
We provide a simulation environment and an open-vocabulary Assembly-by-Disassembly baseline combining language-guided perception, constraint-based planning, and collision-aware execution. The pipeline is evaluated against zero-shot and fine-tuned vision-language-action baselines on 100 tasks per tier. Our system achieves higher success, execution accuracy, and stability across all tiers, especially on long-horizon assemblies.

\noindent\textbf{Project page:} \url{https://workbenchmark.github.io}

%\medbreak
\end{abstract}
\section{Introduction}
Robotic assembly demands precise low-level manipulation and abstract reasoning about the action sequence, while satisfying geometric and stability constraints. 
BlocksWorld~\cite{winograd1972understanding} has long served as a canonical testbed~\cite{slaney2001blocks} for symbolic reasoning over abstract discrete spatial
rearrangement problems, but strips away physics by treating placement as atomic. However, on real robots, this gap between symbolic reasoning and grounded physical execution must be bridged.

\begin{figure}
    \centering
    \begin{tikzpicture}
        \node[anchor=south west, inner sep=0] (initial) at (0,0) {%
            \includegraphics[
                width=6cm,
                height=4.5cm,
                keepaspectratio,
                clip
            ]{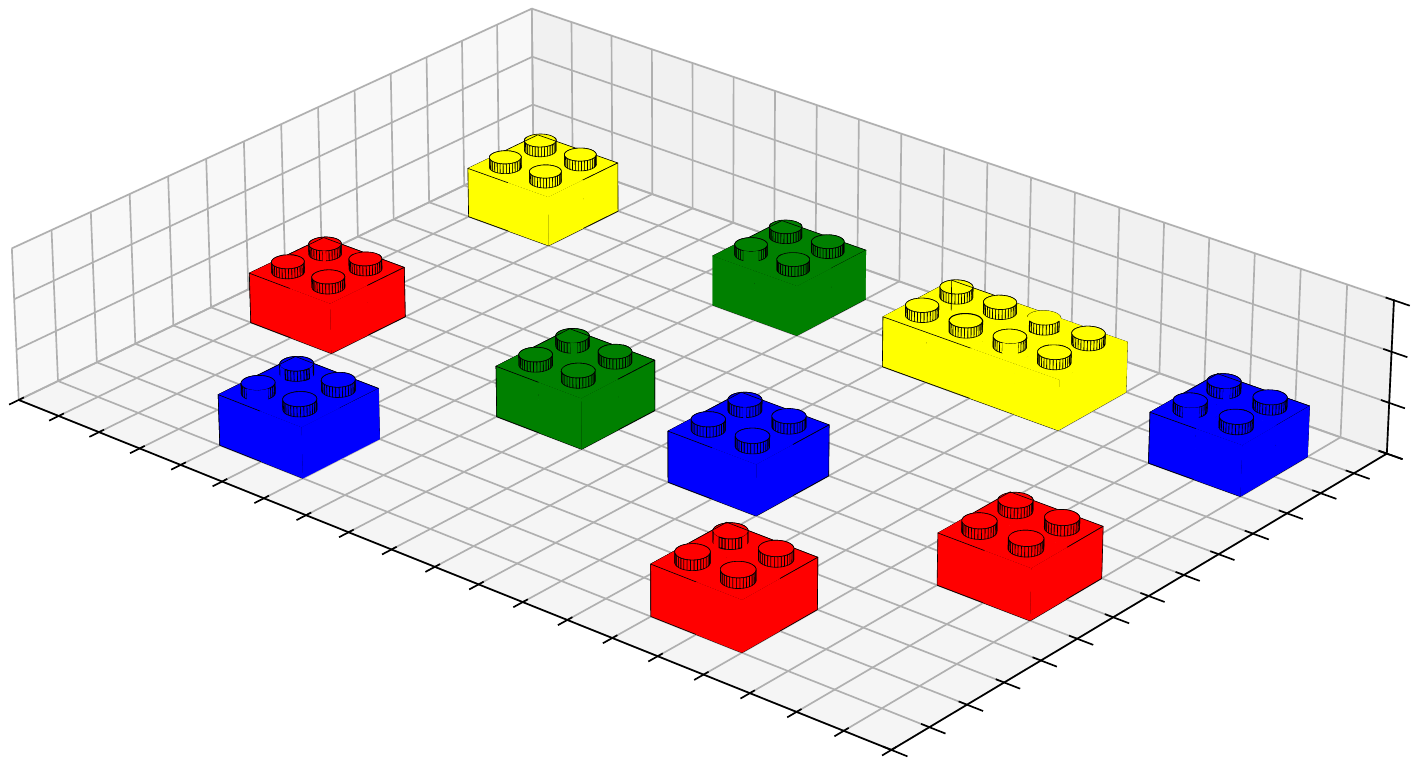}%
        };

        \node[anchor=south west, inner sep=0] (final) at (7,0.5) {%
            \includegraphics[
                width=2.5cm,
                height=4.5cm,
                keepaspectratio,
                clip
            ]{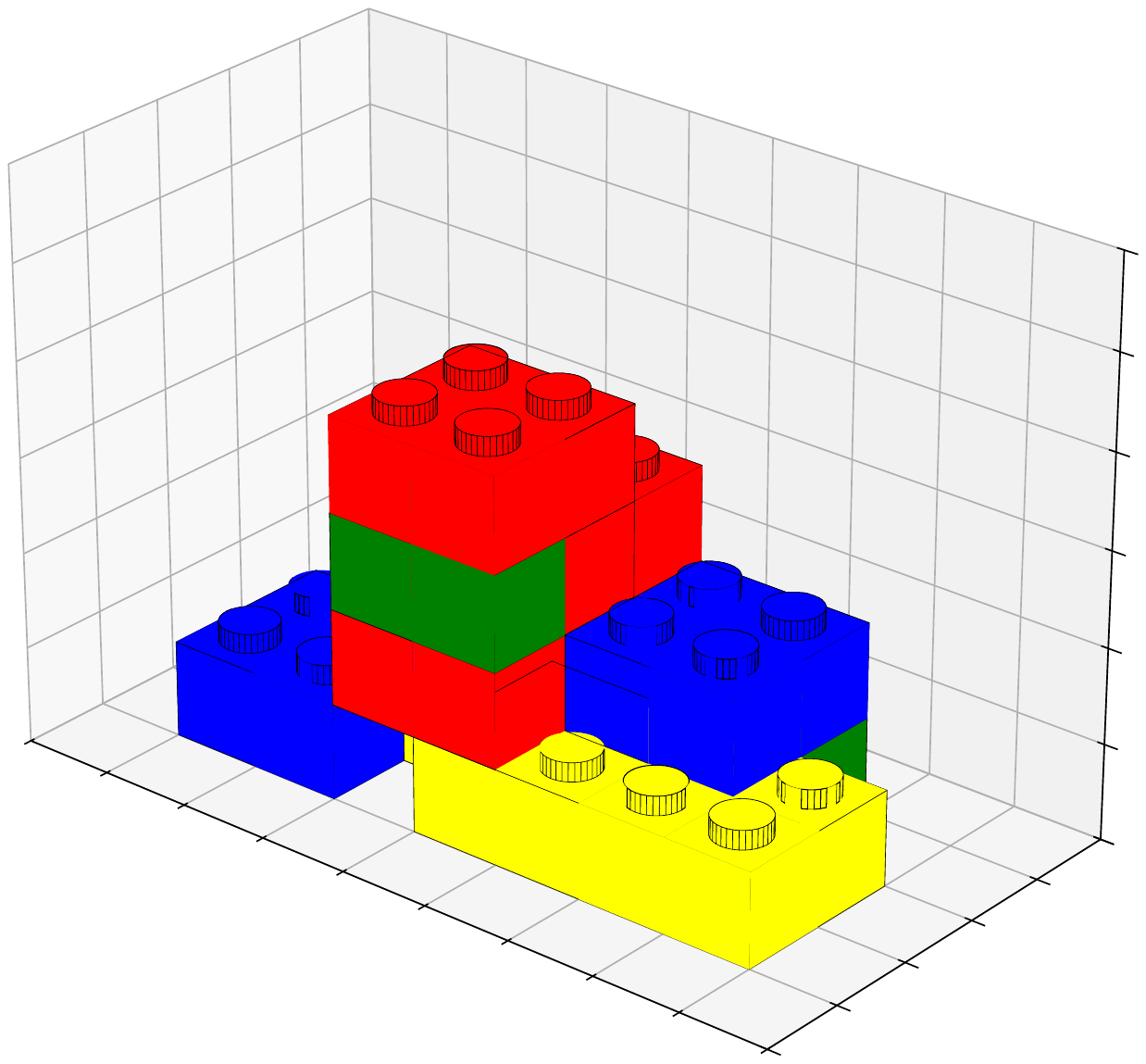}%
        };
        \draw[
            -stealth,
            line width=1.5pt,
            black%yellow
        ] 
        ([xshift=0.12cm]initial.east) -- ([xshift=-0.12cm,yshift=-0.05cm]final.west);
    \end{tikzpicture}
    \caption{Example of an initial state on the pick area (left) and the corresponding final state on the assembly area (right) of a Tier 4 product.}
    \label{fig:init-final}
\end{figure}

\begin{wrapfigure}[21]{R}{0.4\textwidth}
\vspace{-24pt}
\centering
\includegraphics[width=\linewidth]{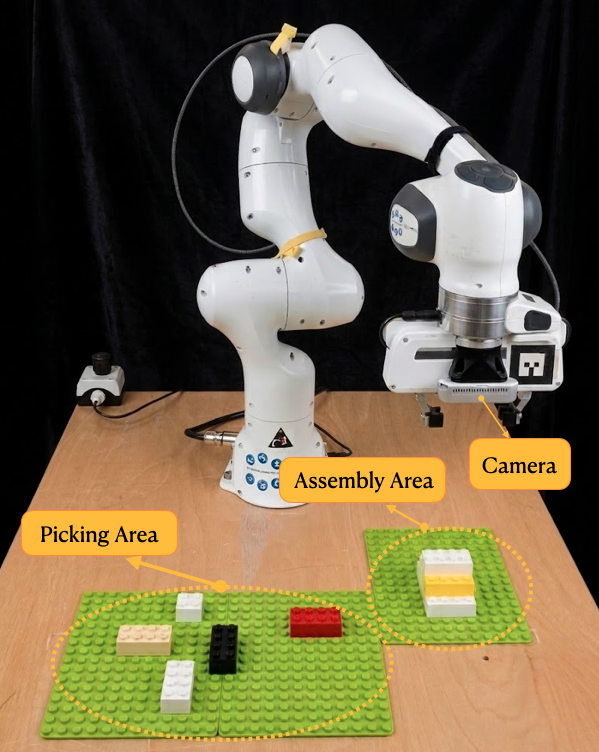} 
\caption{Experimental workspace for the \wbm, including a Franka Emika Panda robot arm, a tabletop assembly area, and LEGO-like bricks.}

\label{fig:workspace}
\end{wrapfigure}

LEGO Duplo bricks are a well-suited medium for assembly tasks as they preserve the core challenges of flexible, small-lot assembly.
Given an initial distribution of parts, a robot needs to reason on multiple levels to achieve a desired target assembly of various Duplo bricks.
The parts are inexpensive, reusable, and mechanically
forgiving, yet small variations in placement have impacts as bricks interlock and block one another. The bricks also allow us to build 
structures whose assembly order is complex. Although LEGO is a recurring robotics testbed~\cite{nagele2020legobot,tian2023asap,kang2021highspeed}, no benchmark compares end-to-end systems.

We introduce \wbm, a
benchmark for end-to-end robotic assembly inspired by autonomous, general assembly in smart manufacturing environments. It targets semi-structured environments with randomized initial configurations and open-world perception, and comprises 400 simulation tasks across four complexity tiers, a real-world evaluation subset, and a simulation environment.
Our task is also reflective of challenges in the recently
established \acf{SML}~\cite{dissanayaka2023rcsml}.

Recent vision-language-action (VLA)
models couple visual grounding with low-level motor
control, guided by a language prompt. They attempt to close this gap by construction: a single model
perceives the scene and generates control output.
We deploy state-of-the-art VLAs on our testbed empirically in
Section~\ref{sec:eval} and find that they struggle with increasing planning horizons. 
To address these shortcomings, we propose an assembly-sequence planning approach using \acfi{ABD}~\cite{homemdemello1991assembly},
which derives an assembly
order by first reasoning about how a completed product can be taken apart. At any state,
only parts that are not blocked are
considered removable, reducing the effective branching factor compared
to forward search. Reversing the resulting disassembly sequence yields
an assembly order that respects precedence relations and avoids
physically infeasible transitions. We
adopt the ABD principle as the planning backbone of our approach, combining it with open-vocabulary object detection,
and motion-planning-based execution.
The contributions of this paper are:
\begin{itemize}
    \item A benchmark formulation inspired by the SML Workbench Track, realised on LEGO Duplo, released with 400 simulation tasks
    across four complexity tiers and an accompanying simulation
    environment.
    \item An open-vocabulary Assembly-by-Disassembly baseline
    integrating constraint-driven planning, open-world perception, and
    collision-aware execution.
    \item An evaluation comparing the baseline against current VLA
    systems in simulation, and a real-world deployment of our approach.
\end{itemize}

\section{Related Work}

We first
position \wbm against existing robotic assembly benchmarks and then provide a review of assembly sequence planning approaches.  

\subsection{Robotic Assembly Benchmarks}
FurnitureBench~\cite{heo2023furniturebench} and the IKEA Assembly
environment~\cite{lee2019ikea} provide reproducible long-horizon manipulation
tasks but focus on furniture-scale objects with largely known initial
configurations.
Kimble et al.~\cite{kimble2020benchmarking} propose evaluation protocols for
small-parts assembly systems, standardizing metrics such as success rate and
cycle time across manipulator platforms.
Lian et al.~\cite{lian2021benchmarking} benchmark off-the-shelf robotic
solutions on NIST Assembly Task Boards, evaluating peg-in-hole insertion across
diverse peg geometries and workspace variations to establish objective baselines.
Their analysis highlights interoperability and scene-awareness as key gaps in
current OTS systems which motivates our open-vocabulary perception approach.
RAMP~\cite{10310102} benchmarks assembly of aluminium extrusion profiles with custom connectors to build large, but flat structures.

Compared with these benchmarks, our setting combines small, modular parts, randomized initial configurations, markerless perception, and the assembly of non-planar structures with overhangs.
Our work is motivated by and situated within the \ac{SML}~\cite{dissanayaka2023rcsml}, which frames assembly as an
end-to-end autonomy challenge in semi-structured environments with randomised
initial part configurations and open-world perception requirements, which
are not jointly addressed by prior benchmarks.
Across robot manipulation research, LEGO bricks have been used
for the very same reasons~\cite{liu2025stablelegostabilityanalysisblock,pun2025generatingphysicallystablebuildable,8593852}. However, no unified benchmark has been established so far, making comparison across approaches difficult.

\subsection{Assembly Sequence Planning}

Nägele et al.~\cite{nagele2020legobot} address coordinated multi-robot LEGO
assembly with a two-layer planner: a symbolic layer orders brick placements via
corridor-based reachability constraints, while a geometric layer generates
collision-free assembly trajectories.
Münker et al.~\cite{munker2023cad} partition CAD-modelled products for automated
disassembly sequence planning via community detection on a graph of part interactions, 
enabling efficient search for feasible sequences. They structure the assembly into sub-assemblies,
thus reducing the number of parts to be considered at any given timestep.
The ASAP framework~\cite{tian2023asap} couples
tree-based search with physics-based simulation to guarantee feasibility at every
intermediate assembly step for complex multi-part structures. In contrast, learning-based approaches, including graph neural
networks~\cite{ma2022graph} and large language models~\cite{valmeekam2023planbench}, can capture high-level
relational patterns but typically lack the strict geometric and physical guarantees
required for interlocking structures.

% In this work, we follow a constraint-driven \ac{ABD} approach, extending it with
% voxel-based grasp reachability and a press-stability criterion specific to the
% stud-engagement insertion model.

\section{The \wbm}

\begin{figure}[t]
    \centering
    % First Row: Tier 1 and Tier 2
    \begin{subfigure}[b]{0.20\textwidth}
        \centering
        \includegraphics[width=\textwidth]{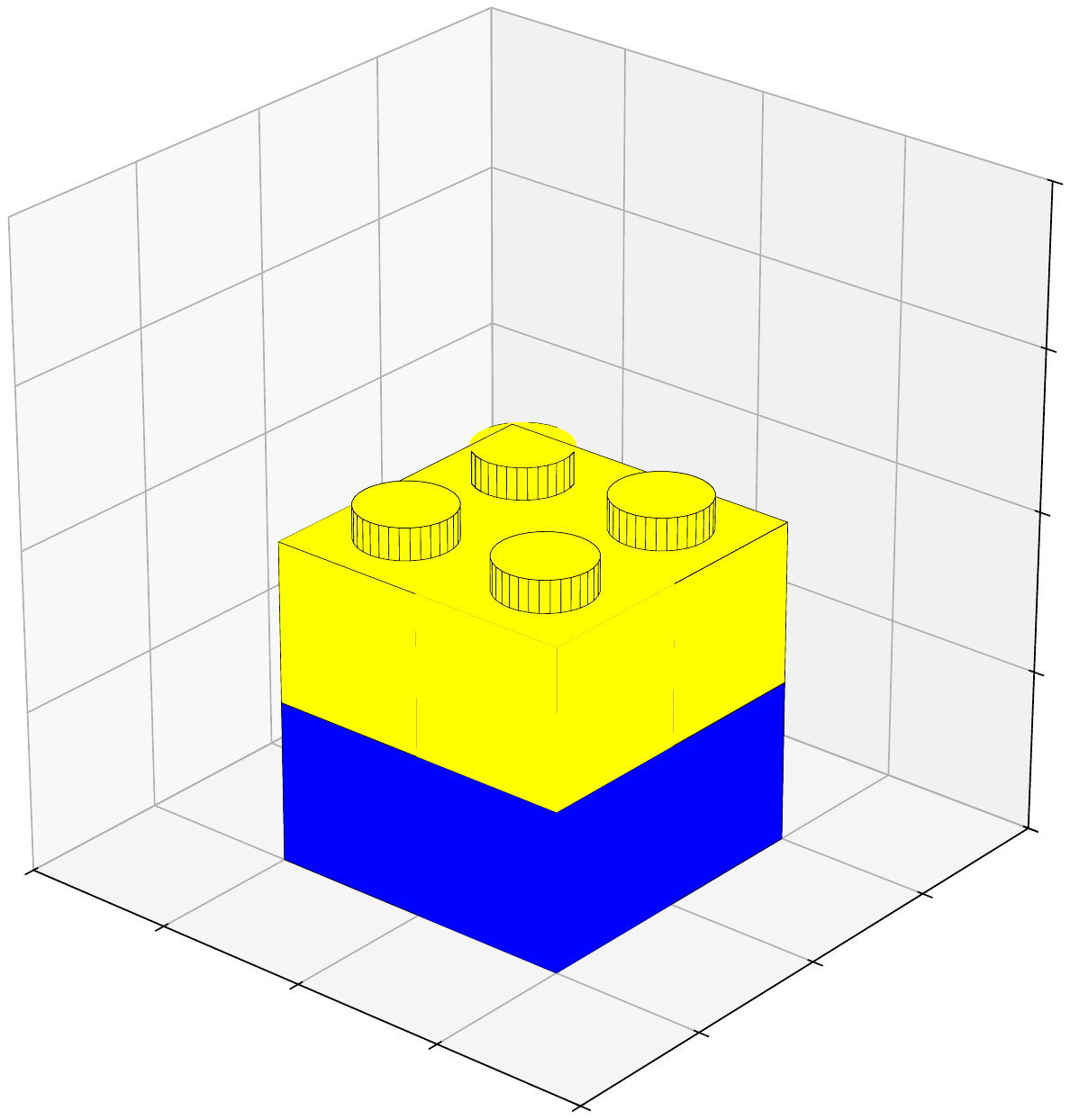}
        \caption{Tier 1}
        \label{fig:tier1}
    \end{subfigure}
    \hfill
    \begin{subfigure}[b]{0.20\textwidth}
        \centering
        \includegraphics[width=\textwidth]{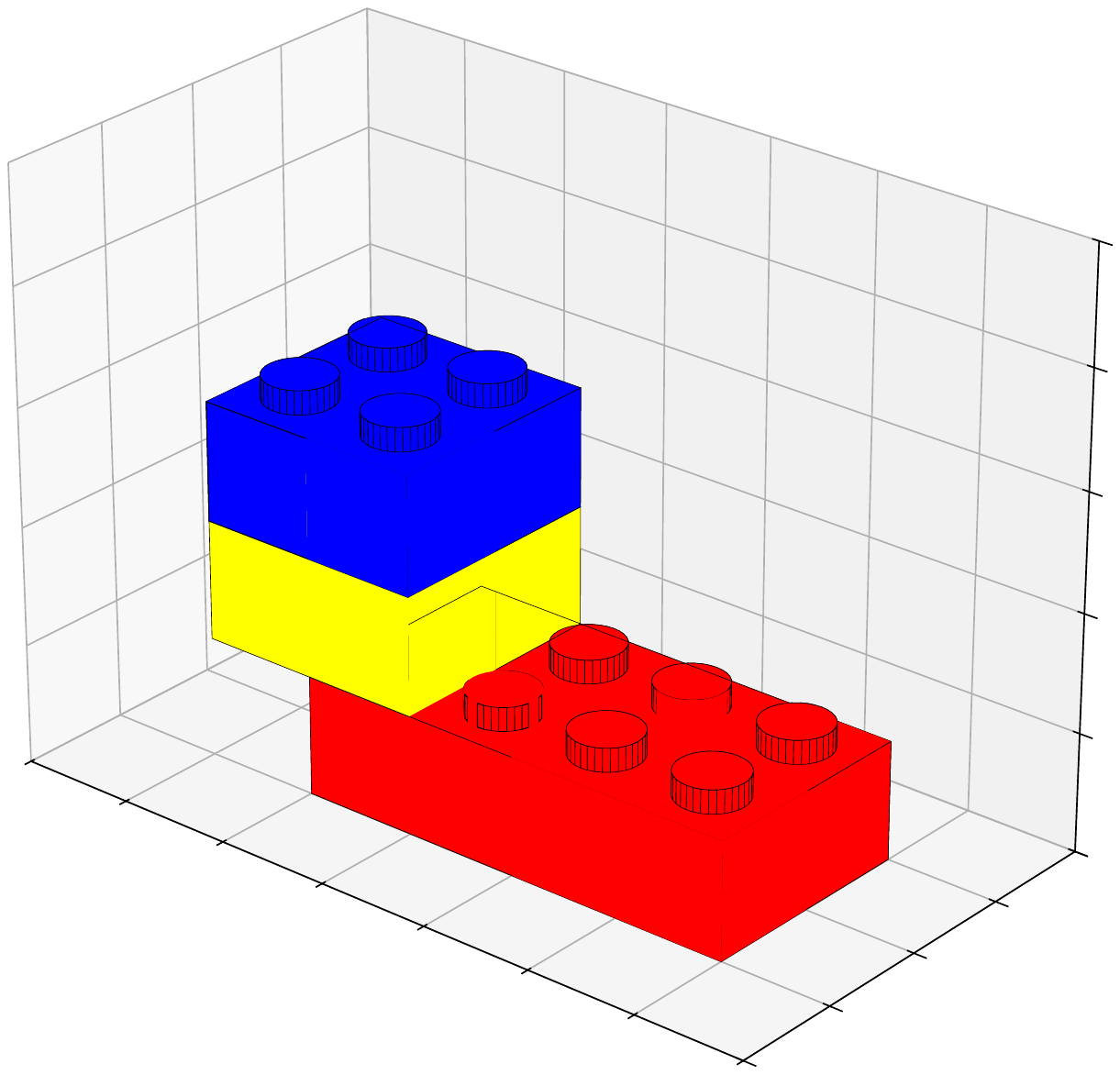}
        \caption{Tier 2}
        \label{fig:tier2}
    \end{subfigure}
    \hfill
    \begin{subfigure}[b]{0.20\textwidth}
        \centering
        \includegraphics[width=\textwidth]{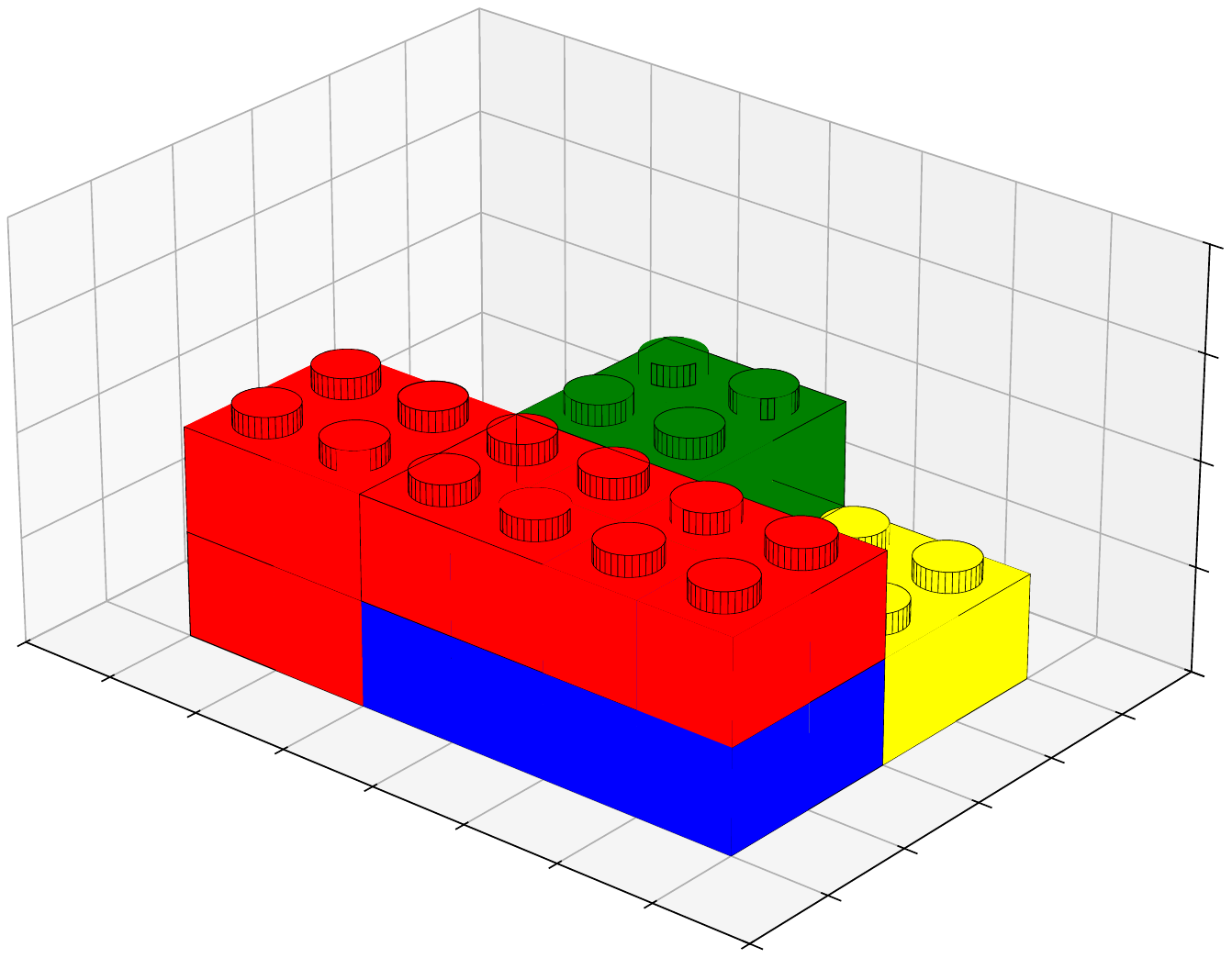}
        \caption{Tier 3}
        \label{fig:tier3}
    \end{subfigure}
    \hfill
    \begin{subfigure}[b]{0.20\textwidth}
        \centering
        \includegraphics[width=\textwidth]{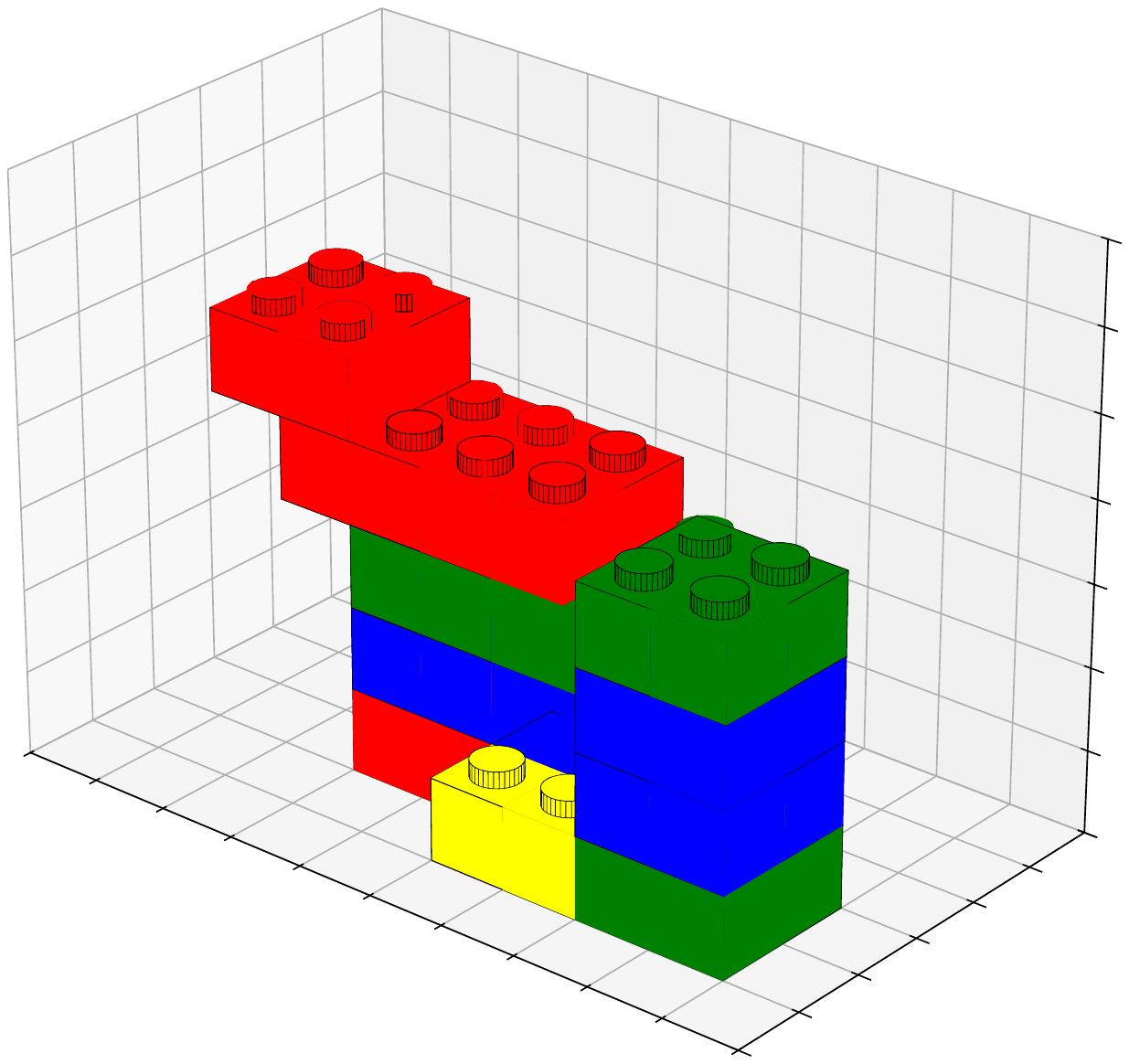}
        \caption{Tier 4}
        \label{fig:tier4}
    \end{subfigure}

    \caption{Examples of the four task complexity tiers in \wbm. From left to right: basic vertical stacking (Tier 1), multi-brick sequential stacking (Tier 2), spatial shape layouts (Tier 3), and complex interlocking structures (Tier 4).}
    \label{fig:complexity_tiers}
\end{figure}

The \wbm consists of 400 LEGO Duplo assembly tasks of varying difficulty organised in tiers. While the stud grid allows for some tolerances, key abilities of small lot, made-to-order manufacturing remain intact: identifying which parts are present, reasoning about a feasible build order for interlocking structures, and placing parts precisely without destabilising the build.
Each task starts in a defined initial state that defines all LEGO Duplo blocks available and their placement in what we call the pick area. The correct blocks need to be moved and assembled into a product defined as the position, rotation, and typing of  blocks (color and shape), using a fully autonomous setup in simulation or in real experiments. We give an example of an initial state and the corresponding final assembly in Figure~\ref{fig:init-final}.

\subsection{Objective}
We formalize our benchmark tasks independently of specific planning
algorithms
and support eventual extensions beyond LEGO.
A task takes place in a workspace comprising two disjoint
regions: a \emph{pick area} $\mathcal{A}_{\text{pick}}$ and an \emph{assembly area}
$\mathcal{A}_{\text{asm}}$.

A target assembly is encoded as a directed graph $\mathcal{G} = (V, E)$, where vertices are bricks labeled by shape and color, and each edge $(u, v, a) \in E$ specifies that brick $v$ is stacked on $u$ with relative pose $a = T_{u,v} \in SE(3)$. Traversing the tree yields target poses $T^*_v \in SE(3)$. Initially, $\mathcal{A}_{\text{pick}}$ contains a flat, non-contacting brick set $\mathcal{P}_0$ with at least 3 cm clearance. $\mathcal{P}_0$ covers all shapes and colors required by $\mathcal{G}$, including multiplicities, and may include distractor blocks.

\begin{wrapfigure}[11]{R}{0.33\textwidth}
\vspace{-2ex}
%\begin{listing}
\setcaptiontype{listing}
\begin{lstlisting}[language=YAML]
  blocks:
      - name:"4x2_brick_1"
        type: "brick_4x2"
        color: yellow
        pos: [0.0, 0.0, 0.0]
        rotation: [0, 0, 0]
      - name: "4x2_brick_2"
        type: "brick_4x2"
        color: blue
        pos: [0.0, 0.0, 0.02]
        rotation: [0, 0, 0]
\end{lstlisting}
\captionsetup{font=scriptsize}
\caption{Example
specification for
a two-brick assembly.}
\label{lst:yaml}
%\end{listing}
\end{wrapfigure}
Both the initial and the target arrangement are described as YAML (Listing~\ref{lst:yaml}), ensuring cross-platform compatibility.
Initial-state information may only be used for task setup, not as a substitute for perception. The robot is required to detect the poses of bricks, plan the assembly, and execute it autonomously without intervention or environment resets. 
A task succeeds if the assembled product is in $\mathcal{A}_{\text{asm}}$ and no longer contacts the robot.

Tasks are categorized into four \removehide{distinct}complexity tiers. Representative examples are provided in Figure~\ref{fig:complexity_tiers}. These tiers independently scale the \textbf{planning challenge} (e.g., order dependencies, interlocking sub\--as\-semblies) and the \textbf{manipulation challenge} (e.g., tight tolerances, multi-layer insertions):
\begin{itemize}
    \item \textbf{Tier 1:} Two-Brick Vertical Stacking (2 bricks)
    \item \textbf{Tier 2:} Multi-Brick Vertical Stacking (3--5 bricks)
    \item \textbf{Tier 3:} 3D Shape Assembly (3--12 bricks)
    \item \textbf{Tier 4:} Complex Shape Assembly (3--12 bricks)
\end{itemize}

Instances are sampled from procedural generators while preserving tier-specific geometric constraints. Tier~1 and Tier~2 use center-aligned vertical stacks with fixed brick type and yaw; Tier~3 samples connected multi-column assemblies with collision-free same-layer placement and at least $50\%$ footprint support for upper bricks; Tier~4 samples assemblies with overhangs, creating order-sensitive support relations. All tasks are verified for plate bounds, scene--goal consistency, initial pick-area clearance, connectivity, same-layer non-overlap, support ratio, and per-layer gripper access. Each tier consists of 100 assemblies, of which 10 are tested for experiments with real robots.

\needspace{5\baselineskip}
\subsection{Setup}
The benchmark is defined for both simulation and real-world execution, with a simulation environment provided. The \wbm requires four main components:
(1) a picking area,
(2) an assembly area,
(3) at least one robot manipulator, and
(4) at least one visual sensor.

An example of this setup is shown in Figure~\ref{fig:workspace}. Custom robotic workbench setups, including one or more manipulators and arbitrary sensing modalities, may be used. However, the system must operate within a constrained footprint of at most $\qtyproduct[product-units = single]{1 x 1 x 2}{\metre}$~(W$\times$D$\times$H).
The size of the operating space is based on the reachability of commonly used robot arms, e.g., UR5 or Franka Research 3, and allows the use of mobile robots such as humanoids.
Both the picking and the assembly area are equipped with LEGO Duplo-compatible base plates that define a discrete grid structure for object placement. The pick area consists of two $24\times24$ LEGO Duplo base plates, while the assembly area consists of a single $24\times24$ base plate.
The picking area serves exclusively as an interface for picking individual bricks. Intermediate assembly operations must only be performed on the assembly area.

\subsection{Metrics}
\label{sec:metrics}

\begin{table}[tb]
    \centering
    \caption{Task-level evaluation metrics.}
    \label{tab:evaluation_metrics}
    \renewcommand{\arraystretch}{1.1}
    \begin{tabular}{@{} p{0.25\linewidth} p{0.72\linewidth} @{}}
        \toprule
        \textbf{Metric} & \textbf{Definition} \\
        \midrule
        Assembly Success &
        If every required brick $v\in V$ is placed at its target pose in the assembly area, attached to the structure.\\
        Execution Accuracy &
        Fraction of brick-to-brick connections whose realised relative pose matches the target. A brick mounted with an offset or not fully pressed down is counted as inaccurate. \\
        Planning Time &
        Wall-clock time spent on perception, reasoning, and motion planning, excluding physical robot motion. \\
        Wall Time &
        Wall-clock duration of solving one instance.\\
        Stability Violation\newline Rate &
        Fraction of bricks in the assembly area that are not connected to the main structure. \\
        \bottomrule
    \end{tabular}
\end{table}

To ensure fair and reproducible comparisons across robotic setups, system performance is evaluated in our benchmark using a standardized set of task-level metrics, reported per tier. The metrics are to be reported as an average over the whole set of problems. We describe our metrics for the benchmark in Table~\ref{tab:evaluation_metrics}.

Success rate measures coarse task completion, whereas execution accuracy is stricter, scoring each connection's translation, yaw, and press-in engagement; a task can thus succeed yet still lose execution accuracy through small local pose errors.

\section{Open-Vocabulary Assembly-by-Disassembly}
We propose a modular system that integrates
constraint-driven planning, open-vocabulary perception, and collision-aware execution.
Our approach realizes the task in three stages:
(1) Planning (2) Perception (3) Execution.

\subsection{Planning: ABD with Voxel-Based Feasibility}
The planning stage determines a sequence of assembly steps to realize the given task.
We represent the current assembly as a discretized voxel state aligned with the LEGO stud grid. Each cell encodes occupancy and brick identity, forming a map $\mathcal{V} \in \mathbb{Z}^{H \times W \times D}$, where $v_{i,j,k}$ is either empty or assigned to a brick. The state is updated after each disassembly step by removing the corresponding cells.

\begin{wrapfigure}[19]{R}{0.5\textwidth}
\vspace{-34pt}
\begin{minipage}{0.5\textwidth}
\begin{algorithm}[H]
\algorithmFontSize
\setcaptiontype{algorithm}
\captionsetup{font=scriptsize}
\caption{ABD reasoning with grasp reachability and press-stability checks}
\label{alg:abd-main}
\begin{algorithmic}[1]
\Function{Disassemble}{$S$}
    \If{$S = \emptyset$}
        \State \Return $\pi$
    \EndIf

    \State $C \gets \Call{TopDownCandidates}{S}$

    \ForAll{$p \in C$}
        \If{$\Call{Reachable}{p,\mathcal{P}(q_0), w, S}$ \textbf{and} $\Call{PressStable}{S \setminus \{p\}}$}
            
            \If{$\Call{IsSubassembly}{p,S}$}
                \State $A \gets \Call{BuildSubassembly}{p,S}$
                \State $\pi_{\text{sub}} \gets \Call{Disassemble}{A}$
                \If{$\pi_{\text{sub}} = \textsc{Fail}$}
                    \State \textbf{continue}
                \EndIf

                \State $\pi_{\text{rest}} \gets \Call{Disassemble}{S \setminus A}$
                \If{$\pi_{\text{rest}} = \textsc{Fail}$}
                    \State \textbf{continue}
                \EndIf

                \State \Return $\pi_{\text{sub}} + \pi_{\text{rest}}$
            \Else
                \State $\pi_{\text{rest}} \gets \Call{Disassemble}{S \setminus \{p\}}$
                \If{$\pi_{\text{rest}} = \textsc{Fail}$}
                    \State \textbf{continue}
                \EndIf

                \State \Return $[p] + \pi_{\text{rest}}$
            \EndIf
        \EndIf
    \EndFor

    \State \Return \textsc{Fail}
\EndFunction
\end{algorithmic}
\end{algorithm}
\end{minipage}
\end{wrapfigure}

Based on the voxel-state representation, we formulate planning as a recursive Assembly-by-Disassembly (ABD) search (Algorithm~\ref{alg:abd-main}). 
The method derives an assembly order by reasoning about how a completed product can be taken apart, which reduces branching compared to forward assembly.
Starting from the target structure, a set of top-down removable candidates $C$ is generated (\textsc{TopDownCandidates}). 
Each candidate $p \in C$ is evaluated using two feasibility checks. 
First, \emph{grasp reachability} (\textsc{Reachable}) requires that at least one predefined grasp pose is collision-free in the current voxel state. 
Second, \emph{press stability} (\textsc{PressStable}) ensures that the remaining structure after removing $p$ can support future insertion. 
We approximate press stability using a support polygon $\mathcal{P}$ defined by ground-contacting bricks and require the projected press location $p_{\mathrm{proj}}$ to lie within $\mathcal{P}$.
If both conditions are satisfied, the algorithm recursively removes $p$. 
If $p$ belongs to a connected sub-assembly, it is processed as a group; otherwise, it is handled individually, as reflected by the recursive calls in Algorithm~\ref{alg:abd-main}. 
If no candidate is valid, the branch is rejected. 
The search proceeds depth-first and returns a disassembly sequence $\pi$, whose reverse defines the final assembly order. 

\begin{figure}[htbp]
\centering

\resizebox{\textwidth}{!}{%
\begin{tikzpicture}[>=latex]

% ===== panel size and spacing =====
\def\panelw{2.85}        % width of each panel
\def\panelh{2.10}        % normal image height
\def\panelhlarge{3.42}   % enlarged height for panels (b) and (d)
\def\gap{0.50}           % gap between panels
\def\arrowy{0.85}        % common arrow height
\def\arrowoffsetl{0.12}  % common arrow offset left
\def\arrowoffsetr{0.12}  % common arrow offset right

% ===== images =====
\node[anchor=south west] (a) at (0,0) {%
    \includegraphics[
        width=\panelw cm,
        height=\panelh cm,
        keepaspectratio,
        clip,
        trim=10 50 20 20
    ]{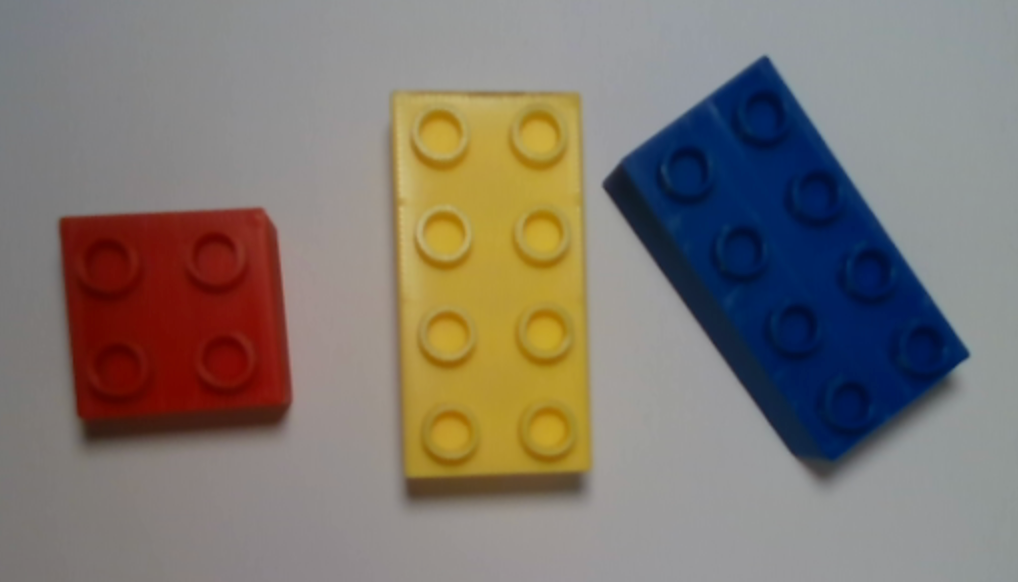}
};

\node[anchor=south west] (b) at (\panelw+\gap,0) {%
    \includegraphics[
        width=\panelw cm,
        height=\panelhlarge cm,
        keepaspectratio,
        clip,
        trim=20 0 80 35
    ]{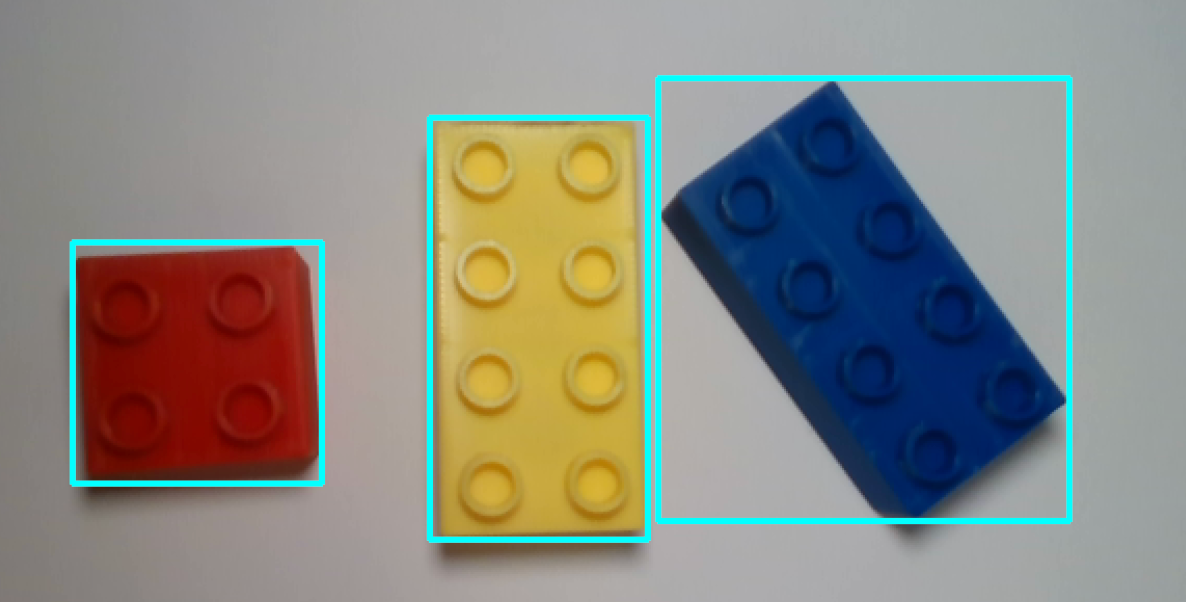}
};

\node[anchor=south west] (c) at (2*\panelw+2*\gap,0) {%
    \includegraphics[
        width=\panelw cm,
        height=\panelh cm,
        keepaspectratio,
        clip,
        trim=0 10 25 55
    ]{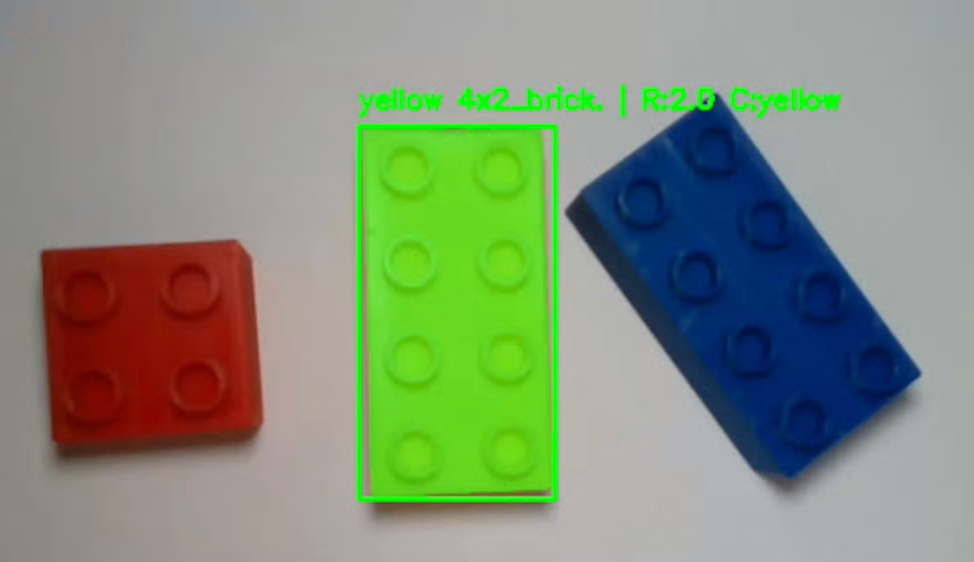}
};

\node[anchor=south west] (d) at (3*\panelw+3*\gap,0) {%
    \includegraphics[
        width=\panelw cm,
        height=\panelhlarge cm,
        keepaspectratio,
        clip,
        trim=35 40 35 40
    ]{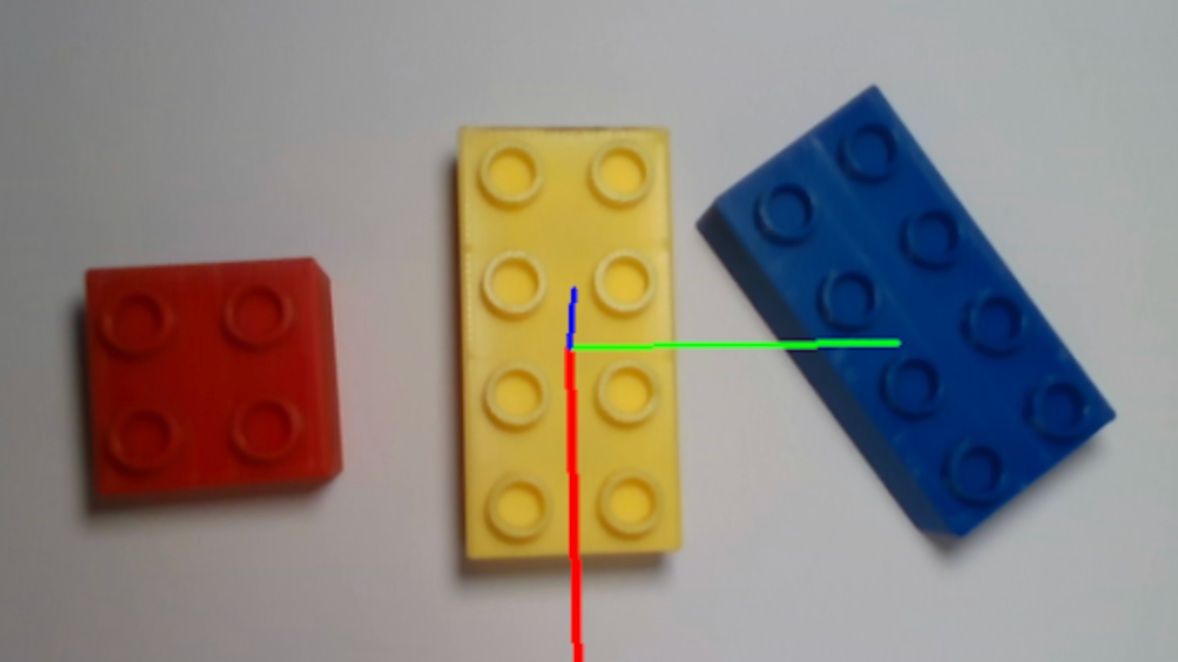}
};

% ===== arrows: same height =====
\draw[->, line width=1.5pt, orange]
(\panelw+\arrowoffsetl,\arrowy) -- (\panelw+\gap+\arrowoffsetr,\arrowy);

\draw[->, line width=1.5pt, orange]
(2*\panelw+\gap+\arrowoffsetl,\arrowy) -- (2*\panelw+2*\gap+\arrowoffsetr,\arrowy);

\draw[->, line width=1.5pt, orange]
(3*\panelw+2*\gap+\arrowoffsetl,\arrowy) -- (3*\panelw+3*\gap+\arrowoffsetr,\arrowy);

% ===== labels: centered under each panel =====
\node[anchor=north, align=center, font=\small]
    at (0.5*\panelw,-0.18)
    {(a) RGB-D\\workspace input};

\node[anchor=north, align=center, font=\small]
    at (\panelw+\gap+0.5*\panelw,-0.18)
    {(b) GroundingDINO\\box};

\node[anchor=north, align=center, font=\small]
    at (2*\panelw+2*\gap+0.5*\panelw,-0.18)
    {(c) SAM\\instance mask};

\node[anchor=north, align=center, font=\small]
    at (3*\panelw+3*\gap+0.5*\panelw,-0.18)
    {(d) FoundationPose\\6D pose};

\end{tikzpicture}
}

\caption{Perception pipeline for active part localisation.}
\label{fig:perception}
\end{figure}

\subsection{Perception: Open-Vocabulary Visual Pose Estimation}
Perception typically relies on fixed-class detectors and known shapes and colors, 
which fail when part identities are specified only at the task level.
We ground detection in free-form text using GroundingDINO~\cite{liu2024groundingdino}. 
Bounding box proposals are refined into segmentation masks using the Segment Anything Model (SAM)~\cite{kirillov2023sam}.
This GroundedSAM
paradigm~\cite{ren2024grounded} enables localisation from
semantic, natural language descriptions without fine-tuning. To get precise 6D poses, we rely on a pre-trained FoundationPose~\cite{wen2024foundationpose}
model that predicts the object poses given the instance mask and ground-truth CAD meshes. Figure~\ref{fig:perception} highlights this flow.

\subsection{Execution: Collision-Aware State Machine}
The execution module implements a state machine that realises each pick-and-place step through the following sequence of actions:
\textsc{Move} (pre-grasp approach) $\to$ \textsc{Open} $\to$
\textsc{Descend} (to grasp) $\to$ \textsc{Grasp} $\to$ \textsc{Lift} $\to$
\textsc{Move} (to target) $\to$ \textsc{Descend} (to place) $\to$ \textsc{Release}.
Transit \textsc{Move} phases are planned as collision-free joint-space trajectories
using MoveIt~2 with the OMPL library~\cite{chitta2012moveit,coleman2014moveit}.
Both \textsc{Descend} phases (grasp and place) are executed as
constrained Cartesian motions to ensure precise vertical alignment during pick
and insertion.
The planning scene models the workspace table and left and right
sideboards as collision objects.
Gripper commands are issued via ROS~2 action clients.
If execution deviates beyond tolerance or a grasp fails, the module retries with
the next available grasp candidate from the planner, before escalating to a
replanning request.

\section{Evaluation}
\label{sec:eval}
We evaluate the proposed system in simulation on our benchmark. The evaluation has three goals: (i) to assess the suitability of the benchmark to provide a tiered measure of complexity of assembly tasks (ii) to assess the performance of the proposed integrated pipeline on structured assembly tasks, and (iii) to compare it against a representative end-to-end vision-language-action (VLA) policy.

\subsection{Experimental Setup}
The experiments are conducted in a LIBERO simulation environment using MuJoCo physics. 
We evaluate both methods on all 100 tasks per tier, sampled from our benchmark. The setup replicates Figure~\ref{fig:workspace}, ensuring that $\mathcal{A}_{\text{asm}}$ and the product configuration are identically populated for all approaches.
All experiments are conducted on a workstation equipped with an NVIDIA RTX 4000 GPU. 
Our approach was also deployed on a real robot setup (Figure~\ref{fig:workspace}), using a Universal Robots UR5 cobot, and a wrist-mounted RGB-D camera, with the planner and controller deployed on the same workstation as in simulation.

\subsection{Baseline Methods}
To provide comparisons, we evaluate two VLA-based baselines.
The first baseline is a zero-shot VLM/VLA pipeline. Starting from the structured product specification, a VLM (Gemini~2.5 Flash~\cite{gemini25flash}) generates a
sequence of assembly instructions. Each instruction specifies the next
brick to pick and its target placement in natural language. These prompts are
then executed by a pretrained $\pi_{0.5}$ VLA\cite{pi0_model} policy.
The second baseline uses the same $\pi_{0.5}$ architecture, but is fine-tuned for \num{18000} gradient steps on demonstrations generated by our ABD pipeline in the same simulator.
The policy observes the agent view and
wrist camera images, together with the robot proprioceptive state, and predicts
closed-loop \texttt{OSC\_POSE} actions. Training and evaluation use the same
simulator, camera setup, state representation, and action space, while the
evaluation tasks are held out from the demonstration set.

We also tested a hybrid variant that uses the assembly plan generated by our
ABD planner as input to the pretrained VLA executor. This did not lead to a
clear improvement over the zero-shot VLM/VLA baseline, suggesting that replacing
the high-level planner alone is insufficient when the VLA executor cannot
reliably follow fine-grained geometric assembly instructions.

\subsection{Evaluation Results}
\begin{table*}[t]
\centering
\caption{
Comparison on the benchmark.
Higher is better for execution accuracy and success rate;
lower is better for planning, wall time, and stability violations. The overall
column reports the unweighted mean over the evaluated tier means.}
\label{tab:tier-metrics}

\renewcommand{\arraystretch}{1.10}
\setlength{\tabcolsep}{4pt}

\begin{tabular}{@{}lccccc@{}}
\toprule
\textbf{Metric}
& \textbf{Tier 1}
& \textbf{Tier 2}
& \textbf{Tier 3}
& \textbf{Tier 4}
& \textbf{Overall} \\
%\midrule
\hline
\rowcolor{gray!20}
\multicolumn{6}{l}{\textbf{Ours (Structured Pipeline)}} \\
Execution Accuracy (\%)   & \textbf{91.60} & \textbf{84.90} & \textbf{62.80} & \textbf{58.40} & \textbf{74.43} \\
Planning Time (s)         & 6.88 & 10.35 & 30.70 & 32.25 & 20.05 \\
Wall Time (s)             & 15.92 & \textbf{23.68} & \textbf{79.08} & \textbf{83.87} & \textbf{50.64} \\
Stability Violations (\%) & \textbf{3.00} & \textbf{5.82} & \textbf{11.37} & \textbf{25.71} & \textbf{11.48} \\
Success Rate (\%)         & \textbf{94.00} & \textbf{87.00} & \textbf{67.00} & \textbf{62.00} & \textbf{77.50} \\

%\midrule
\hline
\rowcolor{gray!20}
\multicolumn{6}{l}{\textbf{Fine-tuned VLA}} \\
Execution Accuracy (\%)   & 74.50 & 57.20 & 22.30 & 10.20 & 41.05 \\
Planning Time (s)         & 2.58 & 2.72 & 3.62 & 4.18 & 3.28 \\
Wall Time (s)             & \textbf{13.15} & 51.97 & 168.64 & 187.52 & 105.32 \\
Stability Violations (\%) & 8.50 & 23.48 & 74.19 & 87.99 & 48.54 \\
Success Rate (\%)         & 82.00 & 63.00 & 23.00 & 2.00 & 42.50 \\

%\midrule
\hline
\rowcolor{gray!20}
\multicolumn{6}{l}{\textbf{VLM/VLA Baseline (Zero-shot)}} \\
Execution Accuracy (\%)   & 62.40 & 51.50 & 15.70 & 12.90 & 35.63 \\
Planning Time (s)         & \textbf{2.24} & \textbf{2.41} & \textbf{3.27} & \textbf{3.78} & \textbf{2.93} \\
Wall Time (s)             & 19.74 & 53.71 & 171.28 & 183.67 & 107.10 \\
Stability Violations (\%) & 36.50 & 56.75 & 83.23 & 88.80 & 66.32 \\
Success Rate (\%)         & 70.00 & 59.00 & 19.00 & 5.00 & 38.25 \\

%\midrule
\hline
\rowcolor{gray!20}
\multicolumn{6}{l}{\textbf{Real Robot Experiments (Structured Pipeline + UR5)}} \\
Wall Time (s)             & 55.77 & 167.98 & 183.33 & -- & 135.69 \\
Success Rate (\%)         & 90.00 & 90.00 & 70.00 & -- & 83.33 \\

\bottomrule
\end{tabular}
\end{table*}

Table~\ref{tab:tier-metrics} summarizes the results on the full simulation
benchmark and the real-robot deployment. Our structured pipeline performs
reliably in the simpler scenarios and maintains more stable behaviour as task
complexity increases. In the lower tiers, the system achieves high success rates
of $94\%$ in Tier~1 and $87\%$ in Tier~2. The transition from Tier~2 to Tier~3
introduces a noticeable increase in both the number of parts and the structural
complexity, which leads to a drop in success rate, though our method still
completes a substantial portion of the tasks.

In contrast, the VLA-based baselines show a sharper degradation as task
complexity increases. Their performance drops in higher tiers, accompanied by
lower execution accuracy, higher wall time, and a higher rate of stability
violations. Although their planning times are shorter, this does not translate
into shorter end-to-end execution. Fine-tuning improves the VLA baseline in the
lower tiers, but it still struggles with complex assemblies.

For the real-robot deployment, we report wall time and success rate on 10 tasks
per tier for Tiers~1--3.
The system achieves $90\%$ success
in Tiers~1 and~2 and $70\%$ in Tier~3, indicating that the benchmark tasks and
our pipeline transfer to a physical setup. Compared with simulation, the
real-robot wall time is higher mainly due to hardware constraints, and
settling time after contact-rich insertions. The remaining failures are mostly
caused by small pose errors accumulated across steps, including hand-eye
calibration error, depth noise, off-center grasps, gripper slip on the plastic
surface, and imperfect press-in alignment.

Overall, the results demonstrate that the proposed structured pipeline provides
more consistent and physically reliable performance across task tiers. Our
asynchronous perception-execution scheme helps reduce wall time by using the
initial visual observation to initialize the scene and by localizing upcoming
bricks in parallel with robot motion.

\section{Conclusion}
We introduced \wbm, a LEGO Duplo product assembly benchmark inspired by the
Workbench Track of the RoboCup Smart Manufacturing League. Our release includes a simulation environment, as well as 400 tasks across 4 complexity tiers. We introduce a solution to this benchmark in the form of an open-vocabulary Assembly-by-Disassembly approach that
integrates constraint-driven ABD planning, open-world perception, and
collision-aware motion planning. Our approach sustains reliable performance in simulation across all four 
complexity tiers while a VLA/VLM-based baseline's performance degrades with spatial complexity. This benchmark is an invitation to robotics manipulation researchers to work on tasks related to the SML, and provides a continuous measure of progress on integrated, real-world assembly.
We plan to scale the benchmark to keep up with the development of robotics manipulation research. First, we will expand the task distribution upward in
complexity, adding disassembly and sub-assembly variants and richer
structural goals beyond the current entry tier. Second, we will move
beyond LEGO Duplo towards heterogeneous industrial parts, narrowing the gap
between benchmark and production setting. We release the benchmark, simulation environment, and baseline implementation to the
community to support this trajectory.

\begin{credits}
\subsubsection{\ackname} The research has been supported by the Alexander von Humboldt Foundation with funds from the German Federal Ministry for Education and Research.
This project was also funded by the German Federal Ministry of Research, Technology and Space (BMFTR) under grant no 02L19C602, and by the BMFTR and the Ministry of Culture and Science of the German State of North Rhine-Westphalia (MKW) under the Excellence Strategy of the Federal Government and the L\"ander.

\subsubsection{\discintname}
The authors have no competing interests to declare that are
relevant to the content of this article.
\end{credits}
\bibliographystyle{splncs04}
\bibliography{bibliography_fixed}

@inproceedings{munker2023cad,
  author    = {M{\"u}nker, S. and Swoboda, D. and {El Zaatari}, K. and Malhotra, N.
               and {Manass{\'e}s Pinheiro de Souza}, L. and G{\"o}ppert, A. M. R.
               and Lee, C.-G. and Schmitt, R. H.},
  title     = {{CAD}-Based Product Partitioning for Automated Disassembly Sequence
               Planning with Community Detection},
  booktitle = {Production Processes and Product Evolution in the Age of Disruption},
  series    = {Lecture Notes in Mechanical Engineering},
  pages     = {570--577},
  year      = {2023},
  publisher = {Springer},
  doi       = {10.1007/978-3-031-34821-1_62}
}

@inproceedings{nagele2020legobot,
  author    = {N{\"a}gele, L. and Hoffmann, A. and Schierl, A. and Reif, W.},
  title     = {{LegoBot}: Automated Planning for Coordinated Multi-Robot Assembly of {LEGO} Structures},
  booktitle = {2020 IEEE/RSJ International Conference on Intelligent Robots and Systems (IROS)},
  pages     = {9088--9095},
  year      = {2020},
  doi       = {10.1109/IROS45743.2020.9341428}
}

@inproceedings{heo2023furniturebench,
  author    = {Heo, M. and Lee, Y. and Lee, D. and Lim, J. J.},
  title     = {{FurnitureBench}: Reproducible Real-World Benchmark for Long-Horizon Complex Manipulation},
  booktitle = {Robotics: Science and Systems (RSS)},
  year      = {2023},
  doi       = {10.15607/RSS.2023.XIX.041}
}

@inproceedings{lee2019ikea,
  author    = {Lee, Y. and Hu, E. S. and Yang, Z. and Yin, A. and Lim, J. J.},
  title     = {{IKEA} Furniture Assembly Environment for Long-Horizon Complex Manipulation Tasks},
  booktitle = {2021 IEEE International Conference on Robotics and Automation (ICRA)},
  pages     = {6343--6349},
  year      = {2021}
}

@inproceedings{liu2024groundingdino,
  author    = {Liu, S. and Zeng, Z. and Ren, T. and Li, F. and Zhang, H. and Yang, J. and Li, C. and Yang, J. and Su, H. and Zhu, J. and Zhang, L.},
  title     = {{Grounding DINO}: Marrying {DINO} with Grounded Pre-Training for Open-Set Object Detection},
  booktitle = {Computer Vision -- ECCV 2024},
  series    = {Lecture Notes in Computer Science},
  pages     = {38--55},
  year      = {2024},
  publisher = {Springer},
  doi       = {10.1007/978-3-031-72970-6_3}
}

@inproceedings{kirillov2023sam,
  author    = {Kirillov, A. and Mintun, E. and Ravi, N. and Mao, H. and Rolland, C. and Gustafson, L. and Xiao, T. and Whitehead, S. and Berg, A. C. and Lo, W.-Y. and Doll{\'a}r, P. and Girshick, R.},
  title     = {Segment Anything},
  booktitle = {2023 IEEE/CVF International Conference on Computer Vision (ICCV)},
  pages     = {4015--4026},
  year      = {2023}
}

@article{kang2021highspeed,
  author  = {Kang, T. and Yi, J.-B. and Song, D. and Yi, S.-J.},
  title   = {High-Speed Autonomous Robotic Assembly Using In-Hand Manipulation and Re-Grasping},
  journal = {Applied Sciences},
  volume  = {11},
  number  = {1},
  pages   = {37},
  year    = {2021},
  doi     = {10.3390/app11010037}
}

@inproceedings{ma2022graph,
  author    = {Ma, L. and Gong, J. and Xu, H. and Chen, H. and Zhao, H. and Huang, W. and Zhou, G.},
  title     = {Planning Assembly Sequence with Graph Transformer},
  booktitle = {2023 IEEE International Conference on Robotics and Automation (ICRA)},
  year      = {2023}
}

@inproceedings{dissanayaka2023rcsml,
  author    = {Dissanayaka, S. and Ferrein, A. and Hofmann, T. and Nakajima, K. and Sanz-Lopez, M. and Savage, J. and Swoboda, D. and Tschesche, M. and Uemura, W. and Viehmann, T. and Yasuda, S.},
  title     = {From Production Logistics to Smart Manufacturing: The Vision for a New RoboCup Industrial League},
  booktitle = {RoboCup 2025: Robot World Cup XXVIII},
  series    = {Lecture Notes in Artificial Intelligence},
  year      = {2025},
  publisher = {Springer}
}

@inproceedings{tian2023asap,
  author    = {Tian, Y. and Willis, K. D. D. and {Al Omari}, B. and Luo, J. and Ma, P. and Li, Y. and Javid, F. and Gu, E. and Jacob, J. and Sueda, S. and Li, H. and Chitta, S. and Matusik, W.},
  title     = {{ASAP}: Automated Sequence Planning for Complex Robotic Assembly with Physical Feasibility},
  booktitle = {2024 IEEE International Conference on Robotics and Automation (ICRA)},
  year      = {2024}
}

@article{ren2024grounded,
  author  = {Ren, T. and Liu, S. and Zeng, A. and Lin, J. and Li, K. and Cao, H. and Chen, J. and Huang, X. and Chen, Y. and Yan, F. and Zeng, Z. and Zhang, H. and Li, F. and Yang, J. and Li, H. and Jiang, Q. and Zhang, L.},
  title   = {Grounded {SAM}: Assembling Open-World Models for Diverse Visual Tasks},
  journal = {arXiv preprint arXiv:2401.14159},
  year    = {2024}
}

@inproceedings{wen2024foundationpose,
  author    = {Wen, B. and Yang, W. and Kautz, J. and Birchfield, S.},
  title     = {{FoundationPose}: Unified 6D Pose Estimation and Tracking of Novel Objects},
  booktitle = {2024 IEEE/CVF Conference on Computer Vision and Pattern Recognition (CVPR)},
  pages     = {17868--17879},
  year      = {2024}
}

@article{chitta2012moveit,
  author = {Chitta, S. and Sucan, I. A. and Cousins, S.},
  title = {{MoveIt!}},
  journal = {IEEE Robotics \& Automation Magazine},
  volume = {19},
  number = {1},
  pages = {18--19},
  year = {2012},
  doi = {10.1109/MRA.2011.2181749}
}

@article{coleman2014moveit,
  author = {Coleman, D. and Sucan, I. A. and Chitta, S. and Correll, N.},
  title = {Reducing the Barrier to Entry of Complex Robotic Software: A {MoveIt!} Case Study},
  journal = {arXiv preprint arXiv:1404.3785},
  year = {2014}
}

@inproceedings{lian2021benchmarking,
  author    = {Lian, W. and Kelch, T. and Holz, D. and Norton, A. and Schaal, S.},
  title     = {Benchmarking Off-The-Shelf Solutions to Robotic Assembly Tasks},
  booktitle = {2021 IEEE/RSJ International Conference on Intelligent Robots and
               Systems (IROS)},
  pages     = {1046--1053},
  year      = {2021},
  doi       = {10.1109/IROS51168.2021.9636586}
}

@article{homemdemello1991assembly,
  author    = {{Homem de Mello}, L. S. and Sanderson, A. C.},
  title     = {A Correct and Complete Algorithm for the Generation of Mechanical
               Assembly Sequences},
  journal   = {IEEE Transactions on Robotics and Automation},
  volume    = {7},
  number    = {2},
  pages     = {228--240},
  year      = {1991}
}

@article{kimble2020benchmarking,
  author = {Kimble, K. and Van Wyk, K. and Falco, J. and Messina, E. R. and Sun, Y. and Shibata, M. and Uemura, W. and Yokokohji, Y.},
  title = {Benchmarking Protocols for Evaluating Small Parts Robotic Assembly Systems},
  journal = {IEEE Robotics and Automation Letters},
  volume = {5},
  number = {2},
  pages = {883--889},
  year = {2020}
}

@misc{pi0_model,
  author       = {{Physical Intelligence}},
  title        = {OpenPI: Open Physical Intelligence Models for Robotics},
  year         = {2024},
  howpublished = {\url{https://github.com/Physical-Intelligence/openpi}},
  note         = {Accessed: 2026}
}

@article{10310102,
  author  = {Collins, J. and Robson, M. and Yamada, J. and Sridharan, M. and Janik, K. and Posner, I.},
  title   = {{RAMP}: A Benchmark for Evaluating Robotic Assembly Manipulation and Planning},
  journal = {IEEE Robotics and Automation Letters},
  year    = {2024},
  volume  = {9},
  number  = {1},
  pages   = {9--16},
  doi     = {10.1109/LRA.2023.3330611}
}

@inproceedings{pun2025generatingphysicallystablebuildable,
  title     = {Generating Physically Stable and Buildable Brick Structures from Text},
  author    = {Pun, Ava and Deng, Kangle and Liu, Ruixuan and Ramanan, Deva and Liu, Changliu and Zhu, Jun-Yan},
  booktitle = {Proceedings of the IEEE/CVF International Conference on Computer Vision (ICCV)},
  year      = {2025}
}

@article{liu2025stablelegostabilityanalysisblock,
  author  = {Liu, Ruixuan and Deng, Kangle and Wang, Ziwei and Liu, Changliu},
  title   = {{StableLego}: Stability Analysis of Block Stacking Assembly},
  journal = {IEEE Robotics and Automation Letters},
  volume  = {9},
  number  = {11},
  pages   = {9383--9390},
  year    = {2024}
}

@inproceedings{8593852,
  author    = {Gilday, K. and Hughes, J. and Iida, F.},
  title     = {Achieving Flexible Assembly Using Autonomous Robotic Systems},
  booktitle = {2018 IEEE/RSJ International Conference on Intelligent Robots and Systems (IROS)},
  pages     = {1--9},
  year      = {2018},
  doi       = {10.1109/IROS.2018.8593852}
}

@article{winograd1972understanding,
  author  = {Winograd, Terry},
  title   = {Understanding Natural Language},
  journal = {Cognitive Psychology},
  volume  = {3},
  number  = {1},
  pages   = {1--191},
  year    = {1972}
}

@article{slaney2001blocks,
  author  = {Slaney, John and Thi{\'e}baux, Sylvie},
  title   = {Blocks World Revisited},
  journal = {Artificial Intelligence},
  volume  = {125},
  number  = {1--2},
  pages   = {119--153},
  year    = {2001}
}

@inproceedings{valmeekam2023planbench,
  author    = {Valmeekam, Karthik and Marquez, Matthew and Olmo, Alberto
               and Sreedharan, Sarath and Kambhampati, Subbarao},
  title     = {{PlanBench}: An Extensible Benchmark for Evaluating Large
               Language Models on Planning and Reasoning about Change},
  booktitle = {Advances in Neural Information Processing Systems 36
               (NeurIPS 2023) Datasets and Benchmarks Track},
  year      = {2023}
}

@misc{gemini25flash,
  author       = {{Google}},
  title        = {{Gemini 2.5 Flash}},
  howpublished = {\url{https://ai.google.dev/gemini-api/docs/models/gemini-2.5-flash}},
  year         = {2026},
  note         = {Google AI for Developers. Last updated: 2026-04-28; accessed: 2026-06-20}
}
\end{document}